\documentclass[10pt,twocolumn,letterpaper]{article}

\usepackage[pagenumbers]{cvpr} 

\usepackage{graphicx}
\usepackage{amsmath}
\usepackage{amssymb}
\usepackage{booktabs}
\usepackage{multirow}

\usepackage{pifont}
\newcommand{\xmark}{\ding{55}}%
\newcommand{\myclubsuit}{$^\text{\ding{168}}$} 
\newcommand{\mydiamondsuit}{$^\text{\ding{169}}$} 
\newcommand{\myspadesuit}{$^\text{\ding{171}}$} 

\usepackage[pagebackref,breaklinks,colorlinks]{hyperref}

\usepackage[capitalize]{cleveref}
\crefname{section}{Sec.}{Secs.}
\Crefname{section}{Section}{Sections}
\Crefname{table}{Table}{Tables}
\crefname{table}{Tab.}{Tabs.}

\def\cvprPaperID{4808} 
\def\confName{CVPR}
\def\confYear{2022}

\usepackage{overpic}
\usepackage{enumitem} 
\usepackage{overpic} 
\usepackage{color}

\definecolor{turquoise}{cmyk}{0.65,0,0.1,0.3}
\definecolor{purple}{rgb}{0.65,0,0.65}
\definecolor{dark_green}{rgb}{0, 0.5, 0}
\definecolor{orange}{rgb}{0.8, 0.6, 0.2}
\definecolor{red}{rgb}{0.8, 0.2, 0.2}
\definecolor{darkred}{rgb}{0.6, 0.1, 0.05}
\definecolor{blueish}{rgb}{0.0, 0.4, .8}
\definecolor{light_gray}{rgb}{0.7, 0.7, .7}
\definecolor{pink}{rgb}{1, 0, 1}
\definecolor{greyblue}{rgb}{0.25, 0.25, 1}

\newcommand{\CIRCLE}[1]{\raisebox{.5pt}{\footnotesize \textcircled{\raisebox{-.6pt}{#1}}}}

\newcommand{\real}{\mathbb{R}}

\newcommand{\balpha}{\bar{\alpha}}

\newcommand{\cent}[1]{\multicolumn{1}{c}{#1}}

\usepackage{blindtext}

\renewcommand{\paragraph}[1]{\vspace{1em}\noindent\textbf{#1}.}
\begin{document}
\title{Dynamic Dual-Output Diffusion Models}

\author{Yaniv Benny\\
Tel Aviv University\\
{\tt\small yanivbenny@mail.tau.ac.il}
\and
Lior Wolf\\
Tel Aviv University\\
{\tt\small wolf@cs.tau.ac.il}
}

\maketitle
\begin{abstract}
Iterative denoising-based generation, also known as denoising diffusion models, has recently been shown to be comparable in quality to other classes of generative models, and even surpass them. Including, in particular, Generative Adversarial Networks, which are currently the state of the art in many sub-tasks of image generation. However, a major drawback of this method is that it requires hundreds of iterations to produce a competitive result. Recent works have proposed solutions that allow for faster generation with fewer iterations, but the image quality gradually deteriorates with increasingly fewer iterations being applied during generation. In this paper, we reveal some of the causes that affect the generation quality of diffusion models, especially when sampling with few iterations, and come up with a simple, yet effective, solution to mitigate them. 
We consider two opposite equations for the iterative denoising, the first predicts the applied noise, and the second predicts the image directly. Our solution takes the two options and learns to dynamically alternate between them through the denoising process.
Our proposed solution is general and can be applied to any existing diffusion model. As we show, when applied to various SOTA architectures, our solution immediately improves their generation quality, with negligible added complexity and parameters. We experiment on multiple datasets and configurations and run an extensive ablation study to support these findings.
\end{abstract}

\section{Introduction}\label{sec:intro}

Over the past few years, deep generative models have reached the ability to generate high-quality samples in various domains, including images~\cite{brock2018large}, speech~\cite{oord2016wavenet}, and natural language~\cite{brown2020language}. For image generation, generative models can be divided into two main branches: approaches based on generative adversarial networks (GAN)~\cite{goodfellow2020generative} and log-likelihood-based methods, such as variational autoencoders (VAE)~\cite{kingma2013auto}, autoregressive models~\cite{oord2016conditional}, and normalizing flows~\cite{rezende2015variational,kingma2018glow}. Log-likelihood models have the advantage of possessing a straightforward objective, which makes them easier to optimize, while GANs are known to be unstable during training~\cite{heusel2017gans, salimans2016improved}. However, until recently, well optimized GAN models outperformed their log-likelihood counterparts in generation quality~\cite{brock2018large, karras2020analyzing, karras2019style, karras2017progressive}.

This changed when Ho et al.~\cite{ho2020denoising} introduced a new type of log-likelihood model called the Denoising Diffusion Probabilistic Model (DDPM). With this model, image quality surpasses GANs~\cite{dhariwal2021diffusion}, while it is also very stable and easy to train. DDPMs follow the concept of iterative denoising: given a noisy image $x_t$, it is gradually denoised by predicting a less noisy image $x_{t-1}$. This process, when done over hundreds (or thousands) of iterations, is able to generate images with very high quality and diversity, even when starting from random noise. DDPMs have many computer vision applications, such as super-resolution~\cite{saharia2021image,li2021srdiff} and image translation~\cite{sasaki2021unit}, and are also extremely effective in non-visual domains~\cite{chen2020wavegrad, luo2021diffusion, rasul2021autoregressive}.

DDPM incorporates a probabilistic denoising process that is dependant on the estimation of the mean component $\mu_{t-1}$. This is done by a neural network parameterized over $\theta$ and denoted as $\mu_\theta(x_t,t)$. However, it was found that through the forward and backward equations this process is better formalized by predicting either the noise $\epsilon_\theta(x_t,t)$ or the original image $x_\theta(x_t,t)$~\cite{ho2020denoising}. Their experiments found the former to be empirically superior, and, as far as we can ascertain, no further comparisons between the two options (noise or original image) have been performed as yet.

In this work, we revisit the original implementation of DDPM, and find that the preference of $\epsilon_\theta$ over $x_\theta$ is circumstantial and depends on the hyperparameters and datasets. In addition, in certain timesteps, the denoising process has less error when predicting the noise component $\epsilon_\theta$, while in others it predicts the original image $x_\theta$ better. 
This realization motivated us to design a model capable of predicting both values and adaptively selecting the more reliable output at each sampling iteration. The modified model has a negligible number of added parameters and complexity. We apply this method to various DDPM models and show a marked improvement in terms of image quality (measured by FID) for many benchmarks. 
This addition to the framework is orthogonal to existing advancements (that we know of), and is able to improve sampling quality, especially with the restriction of few iterations.

\section{Related work}\label{sec:related}
Diffusion probabilistic models were introduced by Sold-Dickstein et al.~\cite{sohl2015deep}, who proposed a model that can learn to reverse a gradual noising schedule. 
This framework is part of long research on generative models that are based on Markov chains~\cite{bengio2014deep,salimans2015markov}, that has led to the development of Noise Conditional Score Networks (NCSN)~\cite{song2019generative,song2020improved} and Denoising Diffusion Probabilistic Models (DDPM)~\cite{ho2020denoising}. Although very similar, DDPMs try to minimize the log-likelihood, while NCSNs optimize the matching objective~\cite{hyvarinen2005estimation}.

The success of DDPMs has sparked a lot of interest in improving upon the original design. Song et al. proposed an implicit sampling (DDIM)~\cite{song2020denoising} that reduces the number of iterations while maintaining high image quality. Nichol and Dhariwal~\cite{nichol2021improved} proposed a cosine noising schedule and a learned denoising variance factor, and in a second work~\cite{dhariwal2021diffusion} proposed architectural improvements and classifier guidance. Watson et al.~\cite{watson2021learning} proposed a dynamic programming algorithm to find an efficient denoising schedule. Nachmani et al.~\cite{nachmani2021denoising} applied a Gamma distribution instead of Gaussian. Luhman and Luhman~\cite{luhman2021knowledge} applied knowledge distillation with DDPMs.

The solution proposed in this work is orthogonal to the contribution of these methods. It is, therefore, possible to apply our method to the above advancements and increase the performance of all these networks. This is demonstrated in our experiments for some of the existing approaches.

\section{Setup}\label{sec:setup}

\begin{figure*}[t]
\centering
\begin{tabular}{@{}c@{}c@{}c@{}}
\includegraphics[width=0.33\textwidth, trim={0 0 0 0}, clip]{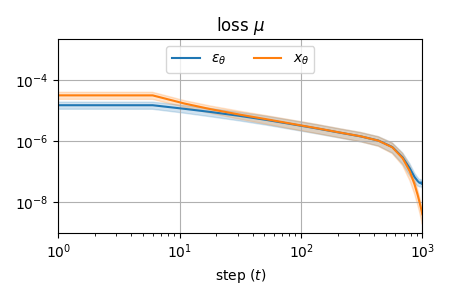} &
\includegraphics[width=0.33\textwidth, trim={0 0 0 0}, clip]{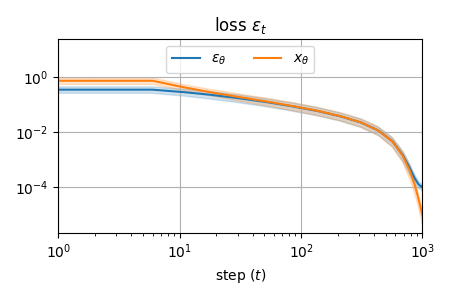} &
\includegraphics[width=0.33\textwidth, trim={0 0 0 0}, clip]{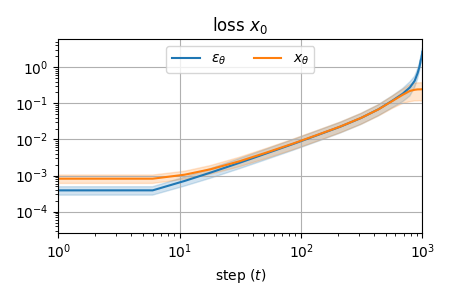} \\
(a)&(b)&(c) \\
\end{tabular}
\caption{{\bf Loss comparison between $\epsilon_\theta$ and $x_\theta$.} (a) Loss on predicting $\tilde\mu_t$, (b) loss on predicting $\epsilon_t$, (c) loss on predicting $x_0$.}
\label{fig:losses}
\end{figure*}

\begin{figure}[t]
\includegraphics[width=\linewidth, trim={10 0 10 0}, clip]{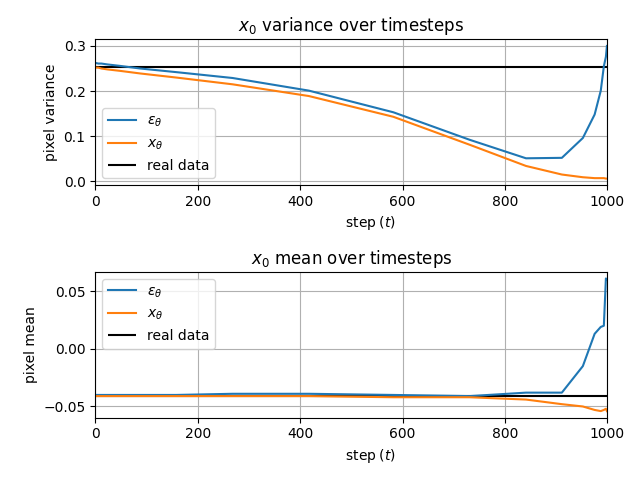}
\caption{Pixel mean and variance of predicted $x_0$ over timesteps, for both the subtractive ($\epsilon_\theta$) and the additive ($x_\theta$) paths. For a model trained on CIFAR10. Real data statistics are in black.}
\label{fig:x0_mean_var}
\end{figure}

Diffusion models operate as the reversal of a gradual noising process. Given a sample $x_0$, we consider the samples $x_t$ for $t \in [1,T]$ obtained by gradually adding noise, starting from $x_0$. Noise is applied in such a way that each instance is noisier than the previous, and the final instance $x_T$ is completely destroyed and can be seen as a sample from a predefined noise distribution. Ho et el.~\cite{ho2020denoising} proposed a Gaussian noise that is applied iteratively as:
\begin{equation}\label{eq:diffusion}
q(x_t | x_{t-1}) := \mathcal{N}(x_t ; \sqrt{1-\beta_t}x_{t-1}, \beta_t  \bf I) 
\end{equation}

Here, $\beta_t \in [0,1]$ for $t \in [1,T]$ are a group of scalars selected so that $x_T \sim \mathcal{N}(x_T; \bf 0, \bf I)$ (a multivariate i.i.d normal distribution). Due to the choice of applying Gaussian noise, a simpler transition can be applied directly from $x_0$ to any $x_t$, which makes training much more efficient. Using $a_t := 1-\beta_t$ and $\balpha_t := \prod_{s=1}^t {a_s}$, we get:
\begin{equation}\label{eq:q_x_t}
    q(x_t | x_0) := \mathcal{N}(x_t; \sqrt{\balpha_t}x_0, (1-\balpha_t) \bf I )
\end{equation}

This formulation also reveals a simpler constraint, which is that $\balpha_T = \prod_{s=1}^T{(1-\beta_s)} \approx 0$, and multiple such schedules have been proposed~\cite{ho2020denoising,nichol2021improved} and tested.

Through this equation, any intermediate step $x_t$ can be sampled, given a noise sample $\epsilon \sim \mathcal{N}(\bf 0, \bf I)$:
\begin{equation}\label{eq:x_t}
    x_t = \sqrt{\balpha_t}x_0 + \sqrt{1-\balpha_t} \epsilon 
\end{equation}

Notice that $x_0$ can easily be backtraced through
\begin{equation}\label{eq:x_t20}
    x_0 = \frac{1}{\sqrt{\balpha_t}}(x_t - \sqrt{1-\balpha_t} \epsilon),
\end{equation}
which is an important property for the denoising process.

The ``reversed'' denoising process is then a Markovian process, parameterized with a neural network over $\theta$ as:
\begin{equation}
    p_\theta(x_{t-1} | x_t) := \mathcal{N}(x_{t-1} ; \mu_\theta(x_t, t), {\bf \Sigma}_\theta(x_t, t))
\end{equation}

Training this model is done by sampling a random $t \in [1,T]$ and minimizing the loss $L_t$ with stochastic gradient descent. $L_t$ is the KL-divergence:
\begin{equation}
    L_t = D_{\text {KL}} (q(x_{t-1} | x_t ) \| p_\theta(x_{t-1} | x_t ))
\end{equation}

Some critical modifications are responsible for the stabilization of this objective. The high variance distribution $q(x_{t-1} | x_t)$ is replaced with the more stable $q(x_{t-1} | x_t, x_0)$, which is practically the combination of the posterior $q(x_{t-1} | x_t)$ and the ``forward'' process $q(x_{t-1} | x_0)$.
\begin{gather}
    q(x_{t-1} | x_t, x_0) := \mathcal{N} (x_{t-1} ; \tilde{\mu}_t(x_t, x_0, t), \tilde{\beta}_t \bf I) \label{eq:q_x_t-1}\\
     \tilde{\mu}_t(x_t, x_0, t) := \frac{\sqrt{\balpha_{t-1}} \beta_t}{1-\balpha_t}x_0 + \frac{\sqrt{\alpha_t} (1 - \balpha_{t-1})}{1-\balpha_t}x_t \label{eq:mu-tilde}\\
     \tilde{\beta}_t := \frac{1-\balpha_{t-1}}{1-\balpha_t} \beta_t \label{eq:beta-tilde}
\end{gather}

As for $p_\theta(x_{t-1} | x_t )$, Ho et al.~\cite{ho2020denoising} found that fixing ${\bf \Sigma}_t$ to a constant $\sigma_t^2$ makes it easier to optimize an objective that is reduced to predicting the mean vector $\tilde{\mu}_t$ only:
\begin{equation}\label{eq:p_theta}
    p_\theta(x_{t-1} | x_t) := \mathcal{N}(x_{t-1} ; \mu_\theta(x_t, t), \sigma_t^2 \bf I)
\end{equation}

As a corollary, the loss function becomes:
\begin{equation}\label{eq:lt_orig}
    L_t := \frac{1}{2\sigma_t^2} \| \tilde{\mu}_t(x_t, x_0, t) - \mu_\theta(x_t,t) \| ^2
\end{equation}

The constant $\sigma_t$ was selected to be $\beta_t$, even though there is also a theoretical explanation for choosing $\tilde{\beta}_t$ instead.

In addition, the prediction of $\mu_\theta$ directly was replaced by either \CIRCLE{1} the prediction of $x_0$, denoted as $x_\theta$, and computing $\mu_\theta$ (denoted as $\mu_x(x_\theta)$) through Eq.~\ref{eq:mu-tilde}, or \CIRCLE{2} predicting $\epsilon$, denoted as $\epsilon_\theta$, and using Eq.~\ref{eq:x_t20},\ref{eq:mu-tilde} ($\mu_\epsilon(\epsilon_\theta)$).  
\begin{align}
    & \CIRCLE{1} & \mu_x(x_\theta) &:= \frac{\sqrt{\alpha_t} (1 - \balpha_{t-1})}{1-\balpha_t}x_t + \frac{\sqrt{\balpha_{t-1}} \beta_t}{1-\balpha_t}x_\theta \label{eq:mu-x} \\
    & \CIRCLE{2} & \mu_\epsilon(\epsilon_\theta) &:= 
            \frac{1}{\sqrt{\alpha_t}}x_t
            - \frac{1 - \alpha_t}{\sqrt{1-\balpha_t}\sqrt{\alpha_t}} \epsilon_\theta
            \label{eq:mu-eps}
\end{align}

The latter was chosen based on empirical evidence, and by further developing the equations this choice resulted in a new formalization of $L_t$ as:
\begin{equation}
    L_t = M_t \| \epsilon - \epsilon_\theta(\underbrace{\sqrt{\balpha_t}x_0 + \sqrt{1-\balpha_t}\epsilon}_{x_t}, t) \| ^2\,,
\end{equation}
where $M_t$ is a weight that should be equal to $\frac{\beta_t^2}{2\sigma_t^2\alpha_t(1-\balpha_t)}$ for consistency with Eq.~\ref{eq:lt_orig}, but was set to $\bf 1$ for simplicity.

\subsection{Pros and cons for predicting $\epsilon$}\label{sec:adv_eps}

There are multiple justifications for why the backward process should be driven by predicting the noise $\epsilon$. The first is that the noise $\epsilon$ has always zero mean and unit variance, and the model can learn these statistics quite easily. A second is that it gives a residual-like equation, where image $x_0$ is predicted by subtracting the output of the model from the input (Eq.~\ref{eq:x_t20}). This provides the model with the option to preserve the information in the input, by predicting zero noise or multiply it by a small $\sqrt{1-\balpha_t}$. This approach becomes increasingly beneficial towards the end of the denoising process, where the amount of noise becomes small, and only minor modifications are needed.

The main disadvantage of this approach is that after the subtraction of noise from $x_t$, the result is scaled with $\sqrt{\balpha_t}$ (Eq.~\ref{eq:x_t20}), which can be a very small value for some steps (large $t$). This can lead to a very large error even for a small error in $\epsilon_\theta$. This error propagates, since the model is limited to modifying the intermediate states with something that resembles noise, and if previous iterations produced a state $x_{t-1}$ that is not viable, it becomes difficult for the model to correct this path. In such cases, multiple iterations may be required just to revert a previous bad prediction. 

This problem is demonstrated in Fig.~\ref{fig:losses},\ref{fig:x0_mean_var}. Fig.~\ref{fig:losses} shows that loss when using $\epsilon_\theta$ is significantly larger at high $t$ than using a direct $x_\theta$ approach. Note that Fig.~\ref{fig:losses} is in log-scale, and the error of $\epsilon_\theta$ is larger by orders of magnitude for hundreds of steps. Fig.~\ref{fig:x0_mean_var} shows the mean and variance of the predicted $x_0$ through the denoising process. As can be seen, the prediction of $x_0$ using $\epsilon_\theta$ starts with very high bias and variance, and it takes multiple steps to correct this. In contrast, the direct prediction using $x_\theta$ immediately starts with very low bias, and its variance increases monotonically with the real data variance.

\begin{figure*}[t]
\centering
\includegraphics[width=\textwidth, trim={30 345 30 390}, clip]{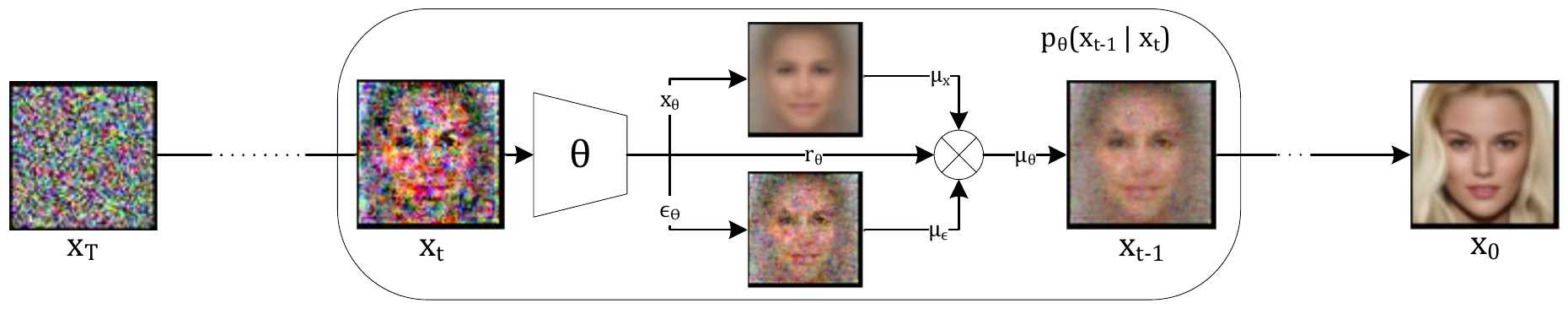}
\caption{{\bf The dual output diffusion model.} A noisy image $x_T$ is gradually denoised into $x_0$. In each iteration, an intermediate state $x_t$ is inserted into a model $f_\theta(x_t,t)$ that predicts $x_\theta, \epsilon_\theta, r_\theta$. These outputs are combined into a mean vector $\mu_\theta$, that is subsequently used to sample the next state $x_{t-1}$.}
\label{fig:arch}
\end{figure*}

\subsection{Pros and cons for predicting $x_0$}\label{sec:adv_x_0}

An advantage and also disadvantage of predicting $x_0$ directly is that the model needs to produce the entire image, not just subtract some noise from its input. This can be an advantage during the first stages when the input image is very noisy. As can be seen in Fig.~\ref{fig:x0_mean_var}, it is easier to predict an unbiased estimate of the image directly than to do so by subtracting noise, and the loss during these steps is substantially lower than $\epsilon_\theta$ (Fig.~\ref{fig:losses}). It becomes a disadvantage during later steps, by when a substantial structure has been formed in $x_t$, and the direct prediction of $x_0$ needs to rebuild it in each step, instead of simply subtracting small noise artifacts. Fig~\ref{fig:losses} shows that all predictions are less accurate for small $t$ when using $x_\theta$.

At first glance, it appears that the backward process using $x_\theta$ loses the residual-like property of $\epsilon_\theta$, since the image $x_0$ is estimated directly and not subtracted from $x_t$. However, the objective of step $t$ is not to predict $x_0$, but to obtain $\tilde\mu_t$. According to Eq.~\ref{eq:mu-tilde}, the residual property is still present in this alternative backward process, which relies on $x_\theta$. 

Note that while $\epsilon_\theta$ is subtracted from $x_t$ (Eq.~\ref{eq:mu-eps}), $x_\theta$ is added to it (Eq.~\ref{eq:mu-x}). 
Therefore, we distinguish between the processes by calling the $\epsilon_\theta$ process the \emph{``subtractive''} backward process, and calling the $x_\theta$ process the \emph{``additive''} backward process.

\section{Method}\label{sec:method}

To leverage the advantages of both flows, we can consider two models. The first, $f_\phi(x_t,t)$, predicts $\epsilon_\phi$ (and $x_\phi$ through Eq.~\ref{eq:x_t20}), and the second, $f_\psi(x_t,t)$, predicts $x_\psi$. 
Each of them can estimate its own $\tilde{\mu}_t$ (Eq.~\ref{eq:mu-x},\ref{eq:mu-eps}), but in order to control how much we want to rely on each model's output, we can interpolate between their estimates with an additional parameter $r_t$, as $r_t \cdot \mu_\psi + (1-r_t) \cdot \mu_\phi$. By selecting a different value for $r_t$ for each step $t$, we can control how much influence we want each path to have on each step.

To simplify and generalize this solution, we fuse $f_\phi,f_\psi$ into one model $f_\theta$, and make the interpolation parameter $r_t$ learned as well ($r_\theta$). The generalized model $f_\theta$ computes:
\begin{gather}
    \epsilon_\theta, x_\theta, r_\theta = f_\theta(x_t, t) \\
    \mu_\theta = r_\theta \cdot \mu_x(x_\theta) + (1-r_\theta) \cdot \mu_\epsilon(\epsilon_\theta)
\end{gather}

An illustration can be seen in Fig.~\ref{fig:arch}. The modifications required to go from a model that only predicts $\epsilon_\theta$ to our new model might seem complex, but they are in fact very simple. The only change to the model is in the number of output channels in the last layer. For example, for $x_0,\epsilon \in \real^{H,W,C}$ and $r \in \real^{H,W,1}$, the output of $f_\theta$ changes from $\real^{H,W,C}$ to $\real^{H,W,2\cdot C+1}$. This means that the number of added parameters should be negligible. Complexity and runtime are also unaffected since the computation of $\mu_\theta$ is negligible.

This new model requires a new loss function $L_t$, which optimizes $e_\theta, x_\theta$, and $r_\theta$. We separate this into three components:
\begin{gather}
    L^\epsilon_{t} = \| \epsilon - \epsilon_\theta \| ^2 \label{eq:loss_eps}\\
    L^x_{t} = \| x_0 - x_\theta \| ^2 \label{eq:loss_x}\\
    L^\mu_{t} = \| \tilde\mu_t - \left( r_\theta [\mu_x(x_\theta)]_{\text{sg}} + (1-r_\theta) [\mu_\epsilon(\epsilon_\theta)]_{\text{sg}} \right) \| ^2 \label{eq:loss_mu}\\    
    L_t = \lambda_t^\epsilon L^\epsilon_{t} + \lambda_t^x L^x_{t} + \lambda_t^\mu L^\mu_{t}\,, 
\end{gather}
where $[\cdot]_{\text{sg}}$ denotes ``stop-grad'', which means that inner values are detached and no gradient propagated back from them. The $\lambda$'s are weights that can be applied to each loss, which we kept as $\bf 1$ throughout our experiments. We found that optimizing with these stop-grads results in a much more stable training regime than the alternative of allowing gradients to propagate through $\mu_x, \mu_\epsilon$, because the gradients of $\frac{\partial \mu_x}{{\partial x_\theta}}, \frac{\partial \mu_\epsilon}{\partial \epsilon_\theta}$ are subject to intense rescaling (see Eq.~\ref{eq:mu-x},\ref{eq:mu-eps}).

\subsection{Implicit sampling}

Song et al.~\cite{song2020denoising} proposed an implicit sampling method (DDIM) that is deterministic after generating the first seed $x_T$. 
Since our method only changes how $\mu_\theta$ is estimated, it does not affect the ability to perform implicit sampling. Their generalized formula is as follows:
\begin{gather}
    q(x_t | x_{t-1},x_0) := \mathcal{N}(\mu_{\small \text {I}}, \sigma_t^2 \bf I) \\
    \mu_{\text {I}} := \sqrt{\balpha_{t-1}} x_0 + \sqrt{1 - \balpha_{t-1} - \sigma_t^2} \cdot \epsilon_t
\end{gather}

In \cite{song2020denoising}, $x_0,\epsilon_t$ were estimated with $\hat{x}_0 = x_0(\epsilon_\theta)$(Eq.~\ref{eq:x_t20}), $\hat{\epsilon}_t=\epsilon_\theta$. Our method uses the interpolated estimation of:
\begin{align}
    \mu_{\text {I}x} &= \sqrt{\balpha_{t-1}} x_\theta + \sqrt{1 - \balpha_{t-1} - \sigma_t^2} \cdot \underbrace{\frac{x_t-\sqrt{\balpha_t}x_\theta}{\sqrt{1-\balpha_t}}}_{\hat{\epsilon}_t} \label{eq:mu_i-x} \\
    \mu_{\text {I}\epsilon} &= \underbrace{\frac{x_t - \sqrt{1-\balpha_t}\epsilon_\theta}{\sqrt{\alpha_t}}}_{\sqrt{\balpha_{t-1}}\hat{x}_0} + \sqrt{1 - \balpha_{t-1} - \sigma_t^2} \cdot \epsilon_\theta \label{eq:mu_i-eps}
\end{align}

When $\sigma_t = 0$, deterministic behavior is obtained, and the model is not sensitive to the value of $\sigma_t^2$ of the probabilistic approach that would have been selected in Eq.~\ref{eq:p_theta}.

An additional advantage of DDIM is that it was shown to be able to generate exceptionally well with far {fewer} iterations than the probabilistic sampling of \cite{ho2020denoising}. For these reasons, our experiments will follow mostly this approach.

\section{Experiments}\label{sec:experiments}

We start by showcasing our method's generation results from each independent output $\epsilon_\theta, x_\theta$, and show how the interpolation parameter $r_\theta$ affects the generation process.
We then evaluate our model against existing state-of-the-art methods on multiple datasets, including CIFAR10~\cite{krizhevsky2009cifar}, CelebA~\cite{liu2015faceattributes}, and ImageNet~\cite{deng2009imagenet}, and perform ablation evaluations with each experiment.

\subsection{Dual-Output Denoising}

\begin{figure*}[t]
\centering
\includegraphics[width=\linewidth, trim={0 0 0 40}, clip]{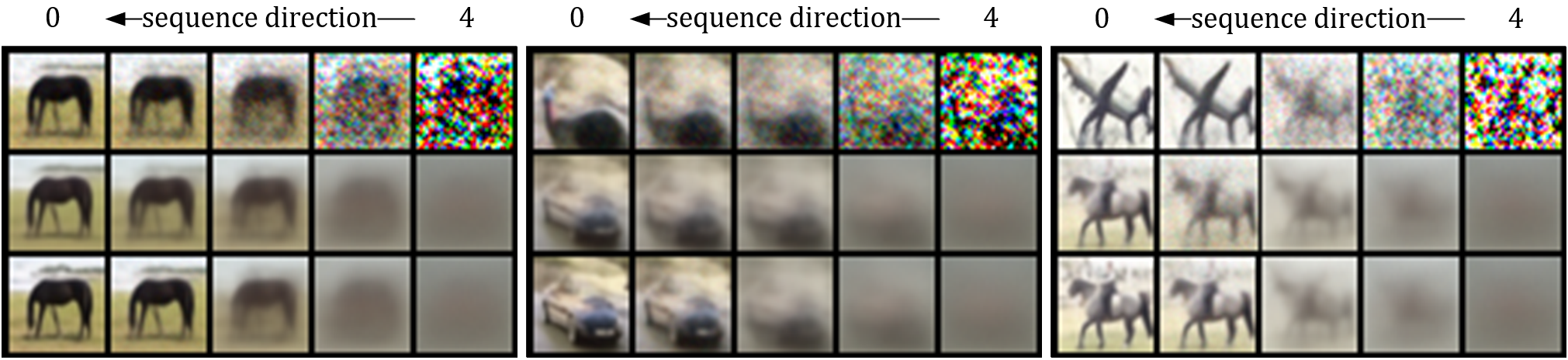} \\
\includegraphics[width=\linewidth, trim={0 0 0 0}, clip]{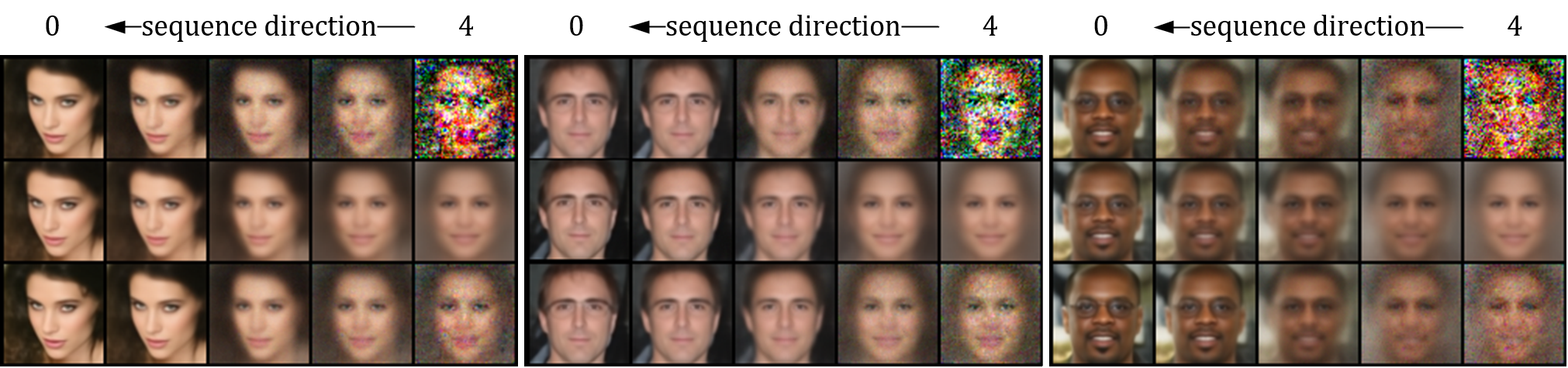} \\
\caption{{\bf Progressive generation in 5 steps.} From top to bottom: \CIRCLE{1} Prediction using $\epsilon_\theta$ (subtractive), \CIRCLE{2} prediction using $x_\theta$ (additive), \CIRCLE{3} our dual-output.  The images generated with the dual-output method are overall cleaner and sharper.}
\label{fig:prog_gen}
\end{figure*}

\begin{figure}[t]
\centering
\includegraphics[width=\linewidth, trim={0 0 0 5}, clip]{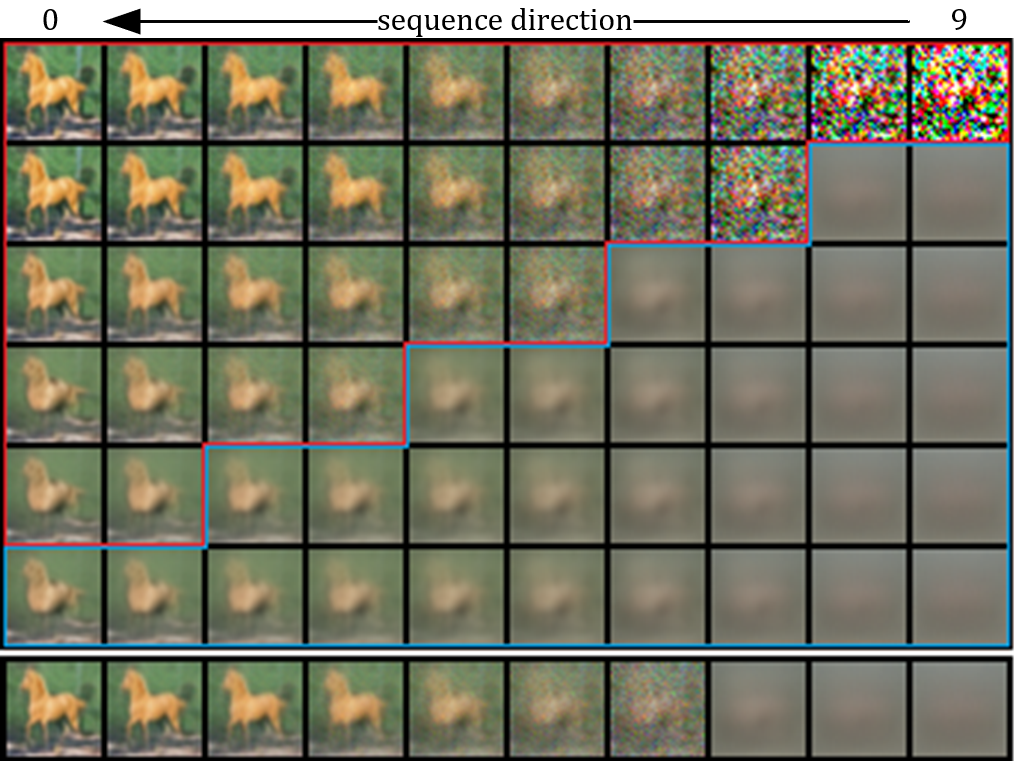} \\
\caption{{\bf Progressive generation in 10 steps.} Each row is a different generation sequence from the same initial noise, with the intermediate results visualized. Steps marked by {\color{red}red} are produced with the $\epsilon_\theta$ output and {\color{blueish}blue} are produced with $x_\theta$. Thus, the sequences in the middle rows start with $\epsilon_\theta$ and switch to $x_\theta$ at some point. 
The sequence at the bottom represents our dynamic dual-output technique.
}
\label{fig:prog_gen_thr}
\end{figure}

\begin{figure}[t]
\includegraphics[width=\linewidth, trim={0 0 0 0}, clip]{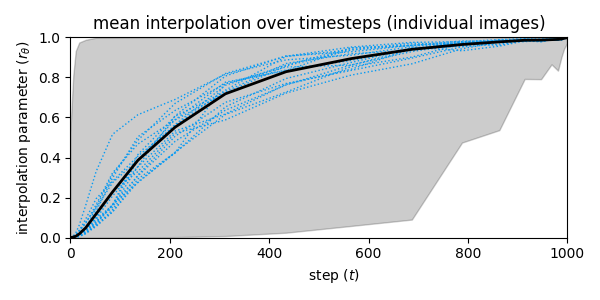} \\
\includegraphics[width=\linewidth, trim={0 0 0 0}, clip]{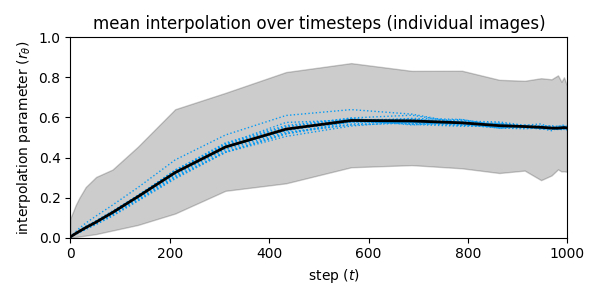}
\caption{Mean value for the interpolation parameter $r_\theta$ over the generation steps. For our model trained on CIFAR10 (top) and CelebA (bottom).}
\label{fig:mean_interpolation}
\end{figure}

As presented in Sec.~\ref{sec:method}, our proposed solution consists of a dual output model. One head predicts the noise $\epsilon_\theta$ while the second predicts the image $x_\theta$, and a third head, $r_\theta$, efficiently balances the two options. To understand how each method affects the iterative process, we visualize their intermediate results during each denoising process. For $x_\theta$, the intermediate result is simply the predicted image. For $\epsilon_\theta$, it is following Eq.~\ref{eq:x_t20}.

These results can be seen in Fig.~\ref{fig:prog_gen}. Evidently, the two outputs produce two very different iterative processes, which to some extent act as opposites. The denoising process that uses $\epsilon_\theta$, produces a very noisy start and gradually removes noise from its previous estimates. In contrast, the process that relies on $x_\theta$ starts with a very blurry image, which resembles an average of many images, and iteratively adds content to it. These two different sequences inevitably result in two different final images, but it appears that the initial seed $x_T$ is a strong enough condition to guide them both in a similar direction in the image space. 

In the bottom row of each grid in Fig.~\ref{fig:prog_gen}, we visualize the dual-output denoising process, driven by the interpolation parameter $r_\theta$. 

Evidently, the interpolation magnitude in each step is dataset dependent. For example, the dual-output process in CIFAR10 starts very similarly to $x_\theta$, while that of CelebA is mixed. It can also be observed that the dual-output result is different from either of the two options.  In CIFAR10, it can be observed that the dual-output produces less noisy images than $\epsilon_\theta$, and sharper than $x_\theta$. For CelebA, we noticed that the dual-output image quality is higher. For example, pieces of hair are more refined and glasses are noticeable.

To better understand the denoising process with each output, we perform an additional experiment, where the method is switched between subtractive and additive at some point in the middle of the process. 
Fig.~\ref{fig:prog_gen_thr} depicts multiple sequences, where the model starts with $x_\theta$ and at some point continues the task with $\epsilon_\theta$ (this order is more natural than starting with $\epsilon_\theta$ and switching to $x_\theta$, see Sec.~\ref{sec:adv_eps}). 
In this figure, the steps surrounded by red boxes are the result of progressing using $\epsilon_\theta$, while steps marked in blue are intermediate results of using $x_\theta$. The top row uses only $\epsilon_\theta$ and the bottom row only $x_\theta$. We again add the sequence produced by the interpolated results. In this example, generation using $x_\theta$ failed to produce a pleasing image, and the other option was superior. However, it seems that some mixture of the two yields the best result. This is the motivation behind our adaptive interpolation, which allows the model to choose dynamically how to proceed.

A valid question would be ``how much better is a dynamic interpolation parameter than learning a constant $r_t$ for each step $t \in [1,T]$?''. To answer this, we measure $r_\theta$ at each step for multiple denoising processes. In Fig.~\ref{fig:mean_interpolation}, we show the average value of $r_\theta$ at each step $t$. The two plots show the average value in black, with the grey region marking the dynamic range of the parameter. We also show the interpolation value for 16 different trajectories in blue. 

This visualization shows that there is a large variability in the trajectory that the interpolation values take. Moreover, when it comes to a particular generation process, our method usually prefers a different value than the overall average. We also evaluate image quality using a fixed $r_t$ in Sec.~\ref{sec:quality}, and compare it to the dynamic $r_\theta$. 

Interestingly, $r_\theta$ seems to behave differently for each dataset. In CIFAR10, interpolation is more clear-cut. It starts with a very high preference towards $x_\theta$, and somewhere around the middle of the process starts to drop fast towards $\epsilon_\theta$.  On CelebA, $r_\theta$ begins at around 0.5 and maintains a relatively narrow dynamic range. Nevertheless, it also drops fast towards $\epsilon_\theta$ in the second half of the process. In both datasets, the model finished with a very high preference for the subtractive process.

\subsection{Image Quality}\label{sec:quality}

\begin{figure*}[t]
\centering
\begin{tabular}{c@{~}c@{~}c}
\includegraphics[width=0.32\textwidth, trim={0 0 0 0}, clip]{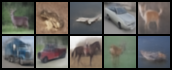} &
\includegraphics[width=0.32\textwidth, trim={0 0 0 0}, clip]{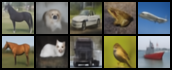} &
\includegraphics[width=0.32\textwidth, trim={0 0 0 0}, clip]{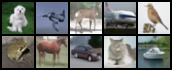} \\
\includegraphics[width=0.32\textwidth, trim={0 0 0 0}, clip]{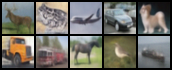} &
\includegraphics[width=0.32\textwidth, trim={0 0 0 0}, clip]{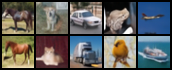} & 
\includegraphics[width=0.32\textwidth, trim={0 0 0 0}, clip]{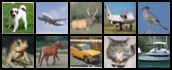}\\
(a)&(b)&(c) \\
\end{tabular}
\caption{{\bf Generation on CIFAR10.} Comparison between similar images. (a) 5, (b) 10, and (c) 20 steps. {\bf Top:} DDIM, {\bf Bottom:} Ours.}
\label{fig:image_compare_cifar}
\end{figure*}

We conduct image quality evaluations on multiple datasets and baselines.
In all evaluations, we use the implicit sampling formula proposed by Song et al.~\cite{song2020denoising}, in order to maintain high quality with low iteration count.
In each evaluation, we specify the number of denoising iterations. 
Timesteps were respaced uniformly, following \cite{song2020denoising}.

We evaluate by generating 50K images for each dataset, which are then compared with the full training set for CIFAR10 and CelebA and the validation set for ImageNet. We evaluate the models on the basis of image quality measured with FID~\cite{heusel2017gans}. FID is known to be sensitive for even slightly different preprocessing and methodology~\cite{parmar2021cleanfid}. For reproducibility and reliability, we use the torch-fidelity~\cite{obukhov2020torchfidelity} library. For ImageNet, we also measure ``improved precision and recall''~\cite{kynkaanniemi2019improved} over VGG feature manifolds between 10K real images and 50K generated images with k=3.

We compare our model to official pretrained models of DDPM~\cite{ho2020denoising}, DDIM~\cite{song2020denoising}, IDDPM~\cite{nichol2021improved}, and ADM~\cite{dhariwal2021diffusion}. We also included IDDPM with implicit sampling (IDDIM), which uses the same official pretrained model, but applies the implicit backward process. Since DDPM and IDDPM enforce a different noising schedule (linear and cosine, respectively), we separate them and compare the models that were trained under equal conditions. DDIM used the pretrained model of DDPM for CIFAR10, but trained a new model for CelebA.

Each comparison to a baseline involves modifying the architecture and loss function to suit our method, and then train the model using the same hyperparameters. Training of each model was performed on 4 NVIDIA RTX 2080 TI GPUS in a distributed fashion. CIFAR10 and CelebA were trained from scratch. On ImageNet, this was not feasible, since the baseline (ADM) took 4.36 million iterations on extremely high-end devices. Instead, we loaded the encoded with the pretrained weights of their model and trained the rest of the model (decoder and residual block) for 80 thousand iterations.

\subsection*{CIFAR10 and CelebA}
In Tab.~\ref{tab:cifar}, we show the results of evaluation on CIFAR10 and CelebA. We perform the evaluations with 5, 10, 20, 50, and 100 iterations. As can be seen, our method outperforms the baselines on all metrics and under all respacing conditions, except for IDDPM with 100 iterations.  The image quality inevitably declines with the reduction of denoising iterations, but our method maintains a significantly lower FID than the equivalent baselines.

When comparing to the ablation experiments, it can be observed that using a fixed $r_t$, which was taken to be the mean value as in Fig.~\ref{fig:mean_interpolation}, worse performance is obtained. The results for the additive and the subtractive paths reveal that no single path is always better than the other. CIFAR10 with linear schedule measured a lower FID with $\epsilon_\theta$, but the cosine schedule and CelebA did better with $x_\theta$.

For a visual comparison, we show generated images of our method alongside DDIM, for 5, 10, and 20 steps, see Fig.~\ref{fig:image_compare_cifar}. 
The images were not cherry-picked, but we did manually select samples in DDIM and our method, that looked relatively similar. For each image in our method, we show the most similar image from 100 generated images in DDIM. 
It can be seen that our method generated better-looking, more detailed and sharper images. 
It is also evident that more steps produce higher quality results.

\begin{table}
\centering
\begin{tabular*}{\linewidth}{@{}l@{~~}l@{~~}l @{\extracolsep{\fill}} r@{~}r@{~}r@{~}r@{~}r@{}}
\toprule
& &                             & \multicolumn{5}{c}{\# iterations} \\
& & Method                      & \cent{5} &  \cent{10} &  \cent{20} &  \cent{50} &  \cent{100} \\
\midrule
\midrule
\multirow{12}{*}{\rotatebox[origin=c]{90}{CIFAR10 (32$\times$32)}} 
& \multirow{6}{*}{\rotatebox[origin=c]{90}{Linear}}
  & DDPM\mydiamondsuit~\cite{ho2020denoising}    
                                & 196.54 & 160.18  & 145.45  & ~65.43  & ~32.65 \\
& & DDIM\mydiamondsuit~\cite{song2020denoising}               
                                & 49.70 & 18.57 & 10.87 & 7.03  & 5.57 \\
& & \textbf{ours}               & \bf 35.12 & \bf 11.68  & \bf 8.62  & \bf 6.68  & \bf 5.54 \\
& & - fixed $r_t$               & 38.50 & 12.08  & 8.71  & 6.89  & 5.57 \\
& & - only $\epsilon_\theta$    & 41.99 & 12.30  & 8.74  & 7.11  & 6.01 \\
& & - only $x_\theta$           & 45.53 & 24.27  & 16.93  & 12.47  & 7.39 \\
        \cmidrule(lr){2-8}
& \multirow{6}{*}{\rotatebox[origin=c]{90}{Cosine}}
  & IDDPM\myclubsuit~\cite{nichol2021improved}    
                                & \cent{\xmark} & 29.10 & 13.33  & 5.73 & \bf 4.58 \\
& & IDDIM\myclubsuit~\cite{nichol2021improved}    
                                & \cent{\xmark} & 38.14 & 19.68 & 8.98  & 6.29 \\
& & \textbf{ours}               & \cent{\xmark} & \bf 18.25 & \bf 12.54 & \bf 5.59 & 5.10 \\
& & - fixed $r_t$               & \cent{\xmark} & 19.60  & 13.93  & 7.24  & 6.17 \\
& & - only $\epsilon_\theta$    & \cent{\xmark} & 36.78 & 17.08 & 8.85  & 6.72 \\
& & - only $x_\theta$           & \bf 45.75 & 19.75 & 13.21 & 7.13  & 5.93 \\
        \midrule
\multirow{6}{*}{\rotatebox[origin=c]{90}{CelebA (64$\times$64)}}
& \multirow{6}{*}{\rotatebox[origin=c]{90}{Linear}}
  & DDPM\myspadesuit~\cite{ho2020denoising}    
                                & 304.89 & 278.31  & 160.67  & 88.74  & 43.90 \\
& & DDIM\myspadesuit~\cite{song2020denoising}    
                                & 56.16 & 16.90  & 13.38  & 8.80  & 6.15 \\
& & \textbf{ours}               & \bf 26.22 & \bf 14.96  & \bf 8.74  & \bf 5.54  & \bf 4.07 \\
& & - fixed $r_t$               & 32.64 & 16.19  & 8.85  & 6.20  & 4.44 \\
& & - only $\epsilon_\theta$    & 64.82 & 27.53  & 12.64  & 9.03  & 8.68 \\
& & - only $x_\theta$           & 29.79 & 16.03  & 9.18  & 6.57  & 4.23 \\
\bottomrule
\multicolumn{8}{c}{Official pretrained models by \cite{ho2020denoising}\mydiamondsuit, \cite{nichol2021improved}\myclubsuit, and \cite{song2020denoising}\myspadesuit.}
\end{tabular*}
\caption{{\bf FID on CIFAR10 and CelebA.} Results are separated by the applied noising schedule ``linear/cosine''. \xmark~marks unstable conditions that produced NaNs; due to dividing by a very low $\balpha$.}
\label{tab:cifar}
\end{table}

\begin{figure*}[t]
\includegraphics[width=\textwidth, trim={0 0 0 0}, clip]{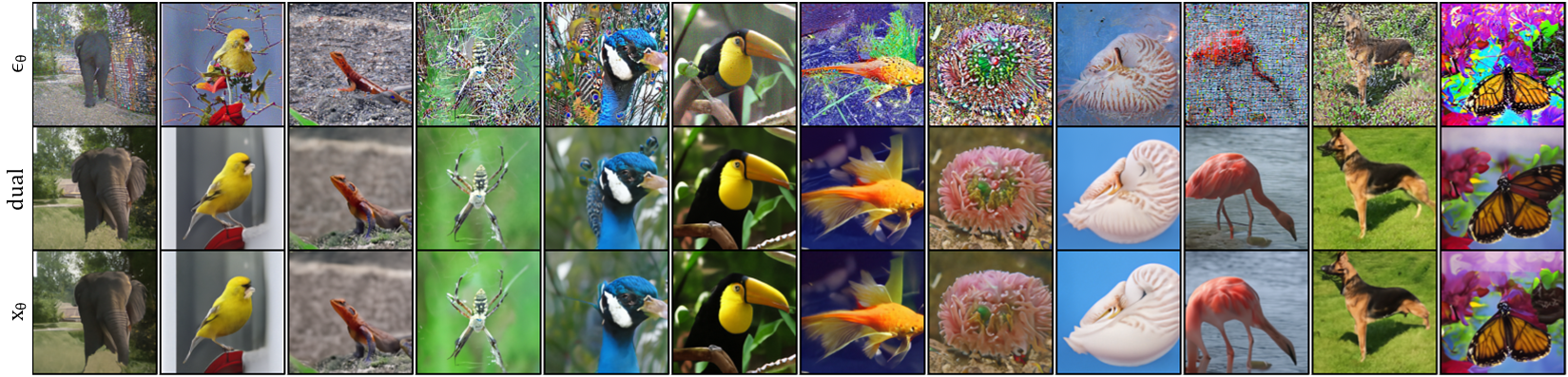}
\caption{{\bf Image generation on ImageNet.} Comparison of generated results for different paths on the same initial noise $x_T$.}
\label{fig:imagenet_gen}
\end{figure*}

\subsection*{ImageNet}

Fig.~\ref{fig:imagenet_gen} shows generated results with our model on ImageNet, for different images generated with the same initial noise $x_T$, but different denoising paths ($\epsilon_\theta$, ``dual'', and $x_\theta$ respectively).
To qualitatively compare the images, we performed a user study, where the subjects were asked to select the most visually convincing image from the three options. Among 25 participants, our images were selected 78\% of the time, followed by 17\% $x_\theta$, and 5\% $\epsilon_\theta$.

Tab.~\ref{tab:imagenet} shows our evaluation on conditional ImageNet with 128$\times$128 resolution. Since we did not train our model for nearly as long as the baseline, we do not compare our results to theirs, but only add them as reference. Our evaluation is focused on comparing the subtractive ($\epsilon_\theta)$ and the additive ($x_\theta$) paths to the dual-output solution. In here, $\epsilon_\theta$ can act as a representative for the baseline, as that is the baseline's method of choice.

The evaluation was performed on 25 and 50 denoising iterations, with and without the classifier guidance proposed by the baseline~\cite{dhariwal2021diffusion}. All evaluations were performed by generating 50 images per class (50K images in total), and comparing them to 50K validation images for FID and 10K images for precision and recall. The results show that the dual-output outperforms each of the alternative paths on all three metrics. Also, again we observed that the results of $x_\theta$ were superior to $\epsilon_\theta$, which shows that the advantage of the subtractive path is circumstantial.
Finally, while there is a considerable gap between our results and the ADM baseline, this evaluation solidifies our speculation about the dual-output process, and suggests that with enough training resources, could surpass the baseline. 

\begin{table}
\centering
\begin{tabular*}{\linewidth}{@{}l@{}r @{\extracolsep{\fill}} @{}c@{}c@{}c@{~}c@{}c@{}c@{}}
\toprule
                            &   & \multicolumn{6}{c}{\# iterations} \\
                            & & \multicolumn{3}{c}{25} & \multicolumn{3}{c}{50} \\
                            \cmidrule(lr){3-5} 
                            \cmidrule(lr){6-8}
\multicolumn{2}{@{}l@{}}{Method (+train steps)}
                            & \cent{FID} & \cent{PR} & \cent{RC} & \cent{FID} & \cent{PR} & \cent{RC} \\
\midrule
\midrule
\multicolumn{8}{c}{ --- no classifier guidance --- } \\
\cmidrule(lr){1-8}
ADM~\cite{dhariwal2021diffusion}
                            & (4.36M) & 11.7 & 0.92 & 0.14 & 7.6 & 0.92 & 0.21 \\
\cmidrule(rl){1-8}
\textbf{dual}               & ($^*$80K) & \bf 27.7 & \bf 0.90 & \bf 0.11 & \bf 25.3 & \bf 0.89 & \bf 0.15 \\
- only $\epsilon_\theta$    & ($^*$80K) & 51.3 & 0.89 & 0.08 & 49.1 & 0.86 & 0.09 \\
- only $x_\theta$           & ($^*$80K) & 29.5 & 0.90 & 0.08 & 27.4 & 0.88 & 0.12 \\
\midrule
\multicolumn{8}{c}{ --- classifier scale 1.0 --- } \\
\cmidrule(lr){1-8}
ADM~\cite{dhariwal2021diffusion}    
                            & (4.36M) & 10.2 & 0.95 & 0.09 & 7.1 & 0.96 & 0.16 \\
\cmidrule(lr){1-8}
\textbf{dual}               & ($^*$80K) & \bf 24.5 & \bf 0.94 & \bf 0.08 & \bf 22.1 & \bf 0.92 & \bf 0.12  \\
- only $\epsilon_\theta$    & ($^*$80K) & 44.1 & 0.93 & 0.07 & 36.8 & 0.89 & 0.07 \\
- only $x_\theta$           & ($^*$80K) & 26.0 & 0.93 & 0.07 & 24.7 & 0.90 & 0.10 \\
\bottomrule
\end{tabular*}
\caption{{\bf Generation evaluation on ImageNet 128$\times$128.} With and without classifier guidance. Measuring FID, precision, and recall. $^*$ Using pretrained encoder from ADM.} 
\label{tab:imagenet}
\end{table}

\section{Discussion and limitations}
While we are able to select an effective value for $r_\theta$ by considering the next-step measure derived from the loss in Eq.~\ref{eq:loss_mu}, this does not necessarily lead to optimal image quality at the end of the generation process. While one can intuitively expect such a greedy approach to be close to optimal, this requires validation. If the greedy approach turns out to be significantly suboptimal, a beam search approach may be able to improve image quality further.

From the societal perspective, the study of diffusion models has two immediate negative outcomes: environmental and harmful use. The environmental footprint of training high-resource neural networks is becoming a major concern. Our work enables the reduction of the number of iterations required to achieve a certain level of visual quality, thus lowering their computation cost. In addition, our experiments are done at a relatively modest energy cost, especially since we opted to train the ImageNet models only in part. The second concern is the ability to generate realistic fake media with generative methods. Our hope is that open academic study of generative models will raise public awareness of the associated risks and enable the development of methods for identifying fake images and audio.

\section{Conclusions}
When applying diffusion models, one can choose to transition to the next step by estimating either the slightly improved image after applying the current step or by estimating the target image.
As we show, the accuracy of each of the two, depends on the exact stage of the inference process. Moreover, the ideal trajectory varies depending on the specific sample, and for most of the process, mixing the two estimates for the next step provides a better results.

\section*{Acknowledgments}
This project has received funding from the European Research Council (ERC) under the European Union's Horizon 2020 research and innovation programme (grant ERC CoG 725974). The contribution of the first author is part of a PhD thesis research conducted at Tel Aviv University.
{
    \small
    \bibliographystyle{ieee_fullname}
    \bibliography{macros,main}
}

\clearpage
\appendix

\section{Model specifications and hyperparameters}

The hyperparameters used across our experiments are the same as the compared baselines, this is in order to perform a concise evaluation.
Still, for clarity and completeness of this work, we indicate the hyperparemeters used for each model version.

\subsection{CIFAR10}
In CIFAR10, we used the baselines DDPM~\cite{ho2020denoising}. The model is a UNet architecture, with the following hyperparameters. The UNet had a depth of 4 downsampling  (and upsampling) blocks, with a base number channel size of 128 and channel multiplier of [1,2,2,2]. Each block contained a residual block with 2 residual layers, and an attention block at the 16x16 resolution. The model was subject to dropout of 0.1. Following the baseline, the linear noise schedule was from 1e-4 to 2e-2 in 1000 steps.
Training was done on 4 GPUs, with a batch size of 128x4, for 1M iterations. The Adam optimizer was used with learning rate of 2e-4, and EMA decay of 0.9999.

For the cosine noise schedule, we used IDDPM~\cite{nichol2021improved}. The improved UNet model included a scale-shift GroupNorm instead of the standard GroupNorm, three residual layers in each block, attention on both the 16x16 and 8x8 resolutions, 4 attention heads instead of 1, and a cosine noise schedule.

\subsection{CelebA}
DDPM was also selected for CelebA evaluation of 64x64 image resolution. The difference from its CIFAR10 counterpart, is the addition of a fifth downsampling layer, with the same base channel size of 64x4, and a channel multiplier of 4.
Model was trained for 500K iterations, using batch size of 32, and Adam optimizer with learning rate 1e-5.

\subsection{ImageNet}
The evaluated model on ImageNet is based on ADM~\cite{dhariwal2021diffusion}. This architecture has some major differences from the previous ones. The model had classifier condition, which were being added to the time condition. Instead of pooled downsampling and interpolated upsampling, a learned up/downsampling was applied through the residual block. In addition, each block has a base channel size of 256, with channel multipliers of [1,1,2,3,4]. Attention of 4 heads was applied on the 32x32, 16x16, and 8x8 resolutions. A dropout of 0.1 and scle-shift GroupNorm. Noise schedule was the default linear schedule.
Finally, we trained only the decoder weights, while using the pretrained weights of the baseline for the rest of the model. Model was trained for 80K iterations, with batch size of 32x4 (with 8 mini-batches of 4). Adam optimizer with learning rate of 1e-5.

\section{Generated images}

In addition to the images in the paper, we provide additional generated images for the various datasets, for further inspection.

\section{Progressive generation}
\begin{figure*}[t]
\centering
\begin{tabular}{@{}cc@{}}
\includegraphics[width=0.48\textwidth, trim={0 0 0 0}, clip]{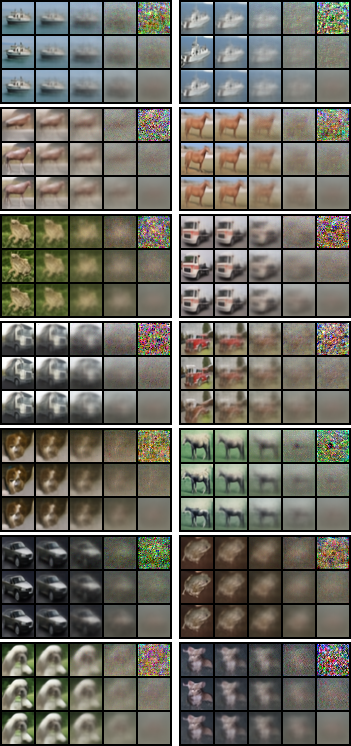} &
\includegraphics[width=0.48\textwidth, trim={0 0 0 0}, clip]{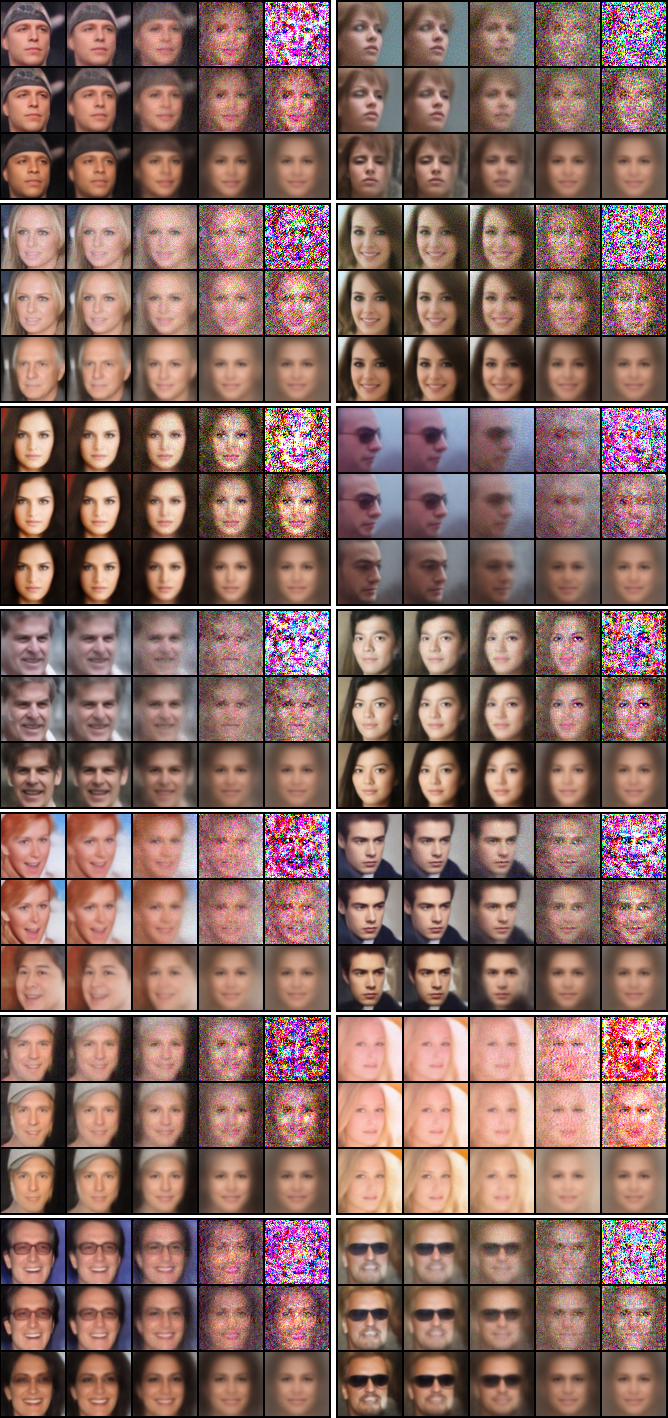} \\
(a)&(b) \\
\end{tabular}
\caption{Progressive generation (a) CIFAR10 and (b) CelebA. From top to bottom: $\epsilon_\theta$, \emph{dual}, and $x_\theta$.}
\label{fig:supp_prog}
\end{figure*}

Fig.~\ref{fig:supp_prog} shows progressive generation results for CIFAR10 and CelebA. All grids show the intermediate results of the three paths $\epsilon_\theta$, \emph{dual}, and $x_\theta$, from top to bottom. It can be seen how in all cases, $\epsilon_\theta$ starts very noisy, while $x_\theta$ is blurry. All paths end with a similar image, but the dual method provides a sharper and less noisy result. In CelebA we noted more difference between the images. The additive path often produced darker images, and the noise in the final result of $\epsilon_\theta$ is very noticeable.

\section{Effect of iteration count}
\begin{figure*}[t]
\centering
\begin{tabular}{@{}cc@{}}
\includegraphics[width=0.47\textwidth, trim={0 0 0 0}, clip]{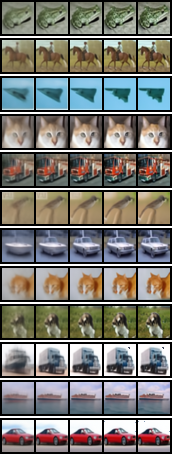} &
\includegraphics[width=0.47\textwidth, trim={0 0 0 0}, clip]{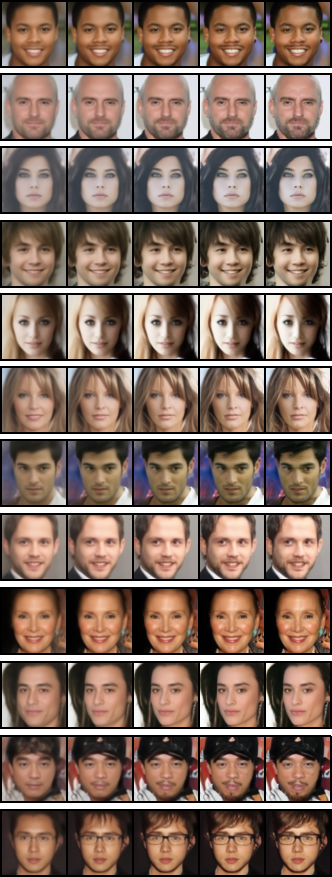} \\
(a)&(b) \\
\end{tabular}
\caption{Generation on (a) CIFAR10 and (b) CelebA. From left to right: 5, 10, 20, 50, 100 iterations.}
\label{fig:supp_generation_t}
\end{figure*}

Fig.~\ref{fig:supp_generation_t} shows generation for both CIFAR10 and CelebA with a different number of denoising iterations. Iteration are monotonically increasing from left to right (5, 10, 20, 50, 100). The effect of the number of iterations is very clear as the image becomes more detailed and sharp when more denoising iterations are applied. Sometimes there is also a change in appearance, but an improvement in quality is always present. However, it can be seen that the change in quality is relatively low, and an already good image is achieved with few iterations.

\section{High quality ImageNet result (128$\times$128)}
\begin{figure*}[t]
\centering
\includegraphics[width=\textwidth, trim={0 0 0 0}, clip]{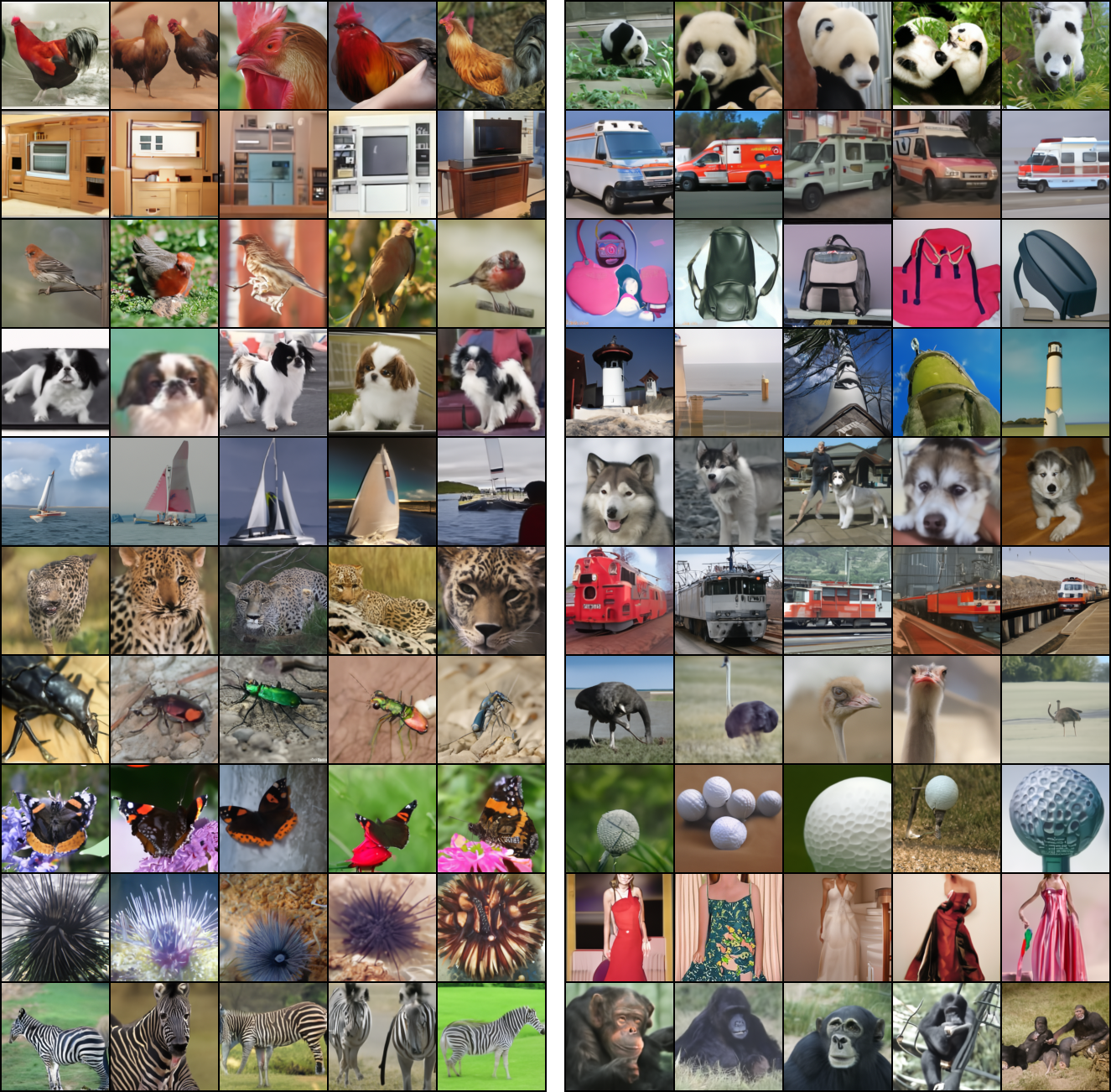}
\caption{Generation on ImageNet with 50 iterations.}
\label{fig:supp_imagenet}
\end{figure*}

Fig.~\ref{fig:supp_imagenet} shows generated images from ImageNet, using 50 denoising iterations. There is a high variety in the images, and the class condition successfully represents the chosen category. Considering that the model was only finetuned for 80K steps, and the denoising is done with only 50 iterations, the image quality is quite good.



\end{document}